\documentclass[10pt,twocolumn,letterpaper]{article}

\usepackage{cvpr}
\usepackage{times}
\usepackage{epsfig}
\usepackage{graphicx}
\usepackage{amsmath}
\usepackage{amssymb}

\usepackage{xspace}
\makeatletter
\DeclareRobustCommand\onedot{\futurelet\@let@token\@onedot}
\def\@onedot{\ifx\@let@token.\else.\null\fi\xspace}

\def\eg{{e.g}\onedot} 
\def\ie{{i.e}\onedot}

\def\etal{\emph{et al}\onedot}
\makeatother

\usepackage{epsfig,graphicx,subfigure}
\usepackage{amsmath,amssymb,dsfont,stmaryrd}
\usepackage{algorithm,algorithmic} 
\usepackage{rotating,multirow}
\usepackage[stable]{footmisc}
\usepackage{theorem}
\usepackage{xcolor}
\usepackage{textcomp}

\usepackage[pagebackref=true,breaklinks=true,letterpaper=true,colorlinks,bookmarks=false]{hyperref}

\cvprfinalcopy 

\def\cvprPaperID{ 597} 

\DeclareMathOperator*{\argmin}{arg\,min} 
\DeclareMathOperator*{\argmax}{arg\,max}

\long\def\ignore#1{}
\def\calW{{\cal W}}
\def\calY{{\cal Y}}

\ifcvprfinal\fi
\begin{document}

\title{A Multi-Plane Block-Coordinate Frank-Wolfe Algorithm \\for Training Structural SVMs with a Costly max-Oracle} 

\author{Neel Shah\\
IST Austria\\
{\tt\small neelshah@ist.ac.at}
\and
Vladimir Kolmogorov\\
IST Austria\\
{\tt\small vnk@ist.ac.at}
\and
Christoph H. Lampert\\
IST Austria\\
{\tt\small chl@ist.ac.at}
}

\maketitle

\begin{abstract}
    Structural support vector machines (SSVMs) are amongst the 
    best performing models for structured computer vision tasks, 
    such as semantic image segmentation or human pose estimation. 
    Training SSVMs, however, is computationally costly, because it 
    requires repeated calls to a structured prediction subroutine 
    (called \emph{max-oracle}), which has to solve an optimization 
    problem itself, \eg a graph cut. 
	
    In this work, we introduce a new algorithm for SSVM training that
    is more efficient than earlier techniques when the max-oracle is 
    computationally expensive, as it is frequently the case in computer
    vision tasks. The main idea is to (i) combine the recent stochastic 
    Block-Coordinate Frank-Wolfe algorithm with efficient hyperplane 
    caching, and (ii) use an automatic selection rule for deciding whether 
    to call the exact max-oracle or to rely on an approximate one based
    on the cached hyperplanes. 
    
    We show experimentally that this strategy leads to faster convergence 
    to the optimum with respect to the number of requires oracle calls, 
    and that this translates into faster convergence with respect to the 
    total runtime when the max-oracle is slow compared to the other steps 
    of the algorithm.

    A publicly available C++ implementation is provided.
\end{abstract}

\section{Introduction}\label{sec:intro}
Many computer vision problems have a natural formulation as 
\emph{structured prediction tasks}: given an input image 
the goal is to predict a structured output object, for 
example a segmentation mask or a human pose. 
Structural support vector machines (SSVMs)~\cite{taskar,tsochan},
are currently one of the most popular methods for learning 
models that can perform this task from training data. 
In contrast to ordinary support vector machines (SVMs)~\cite{VapnikCortesSVM},
which only predict single values, \eg a class label, SSVMs are designed
such that, in principle, they can predict arbitrary structured objects. 
However, this flexibility comes at a cost: training an SSVM requires 
solving a more difficult optimization problem than training an 
ordinary SVM.
In particular, SSVM training requires repeated runs of the structured 
prediction step (the so called \emph{max-oracle}) across the training 
set. 
Each of these steps is an optimization problem itself, \eg finding 
the minimum energy labeling of a graph, 
and often computationally costly. 
In fact, the more challenging the problem is, the more the max-oracle 
calls become a computational bottleneck. 
This is also a major factor why SSVMs are typically only used for 
problems with small and medium-sized training sets, not for large scale 
training as it is common these days, for example, in object 
categorization~\cite{imagenet_cvpr09}. 

In this work, we introduce a new variant of the Frank-Wolfe  
algorithm that is specifically designed for training SSVMs in 
situations where the calls to the max-oracle are the computational 
bottleneck. 
It extends the recently proposed block-coordinate Frank-Wolfe (BCFW)
algorithm~\cite{BCFW} by introducing a caching mechanism that keeps 
the results of earlier oracle calls in memory. 
In each step of the optimization, the algorithm decides whether to 
call the exact max-oracle, or to reuse one of the results from 
the cache. 
The first option might allow the algorithm to make a larger steps 
towards the optimum, but it is slow. The second option will make 
smaller steps, but this might be justified, since every step will 
be much faster. 
Overall, a trade-off between both options will be optimal, and a second 
contribution of the manuscript is a geometrically motivated criterion 
for dynamically deciding at any time during the runtime of the algorithm,
which choice is more promising.

We report on experiments on four different datasets that 
reflect a range of structured prediction scenarios:
multiclass classification, sequence labeling, figure-ground
segmentation and semantic image segmentation.


\section{Structural Support Vector Machines}\label{sec:related}
The task of \emph{structured prediction} is to predict structured objects, 
$y\in\mathcal{Y}$, for given inputs, $x\in\mathcal{X}$. 
Structural support vector machines (SSVMs)~\cite{taskar,tsochan} offer a principled 
way for learning a structured prediction function, $h:\mathcal{X}\to\mathcal{Y}$, 
from training data in a maximum margin framework.
%
%
We parameterize $h(x) = \argmax_{y \in \mathcal{Y}} \langle w, \phi(x,y)\rangle$,
where $\phi:\mathcal{X}\times \mathcal{Y} \rightarrow \mathbb{R}^d$ is a joint feature 
function of inputs and outputs, and $\langle\cdot,\,\cdot\rangle$ denotes the inner
product in $\mathbb{R}^d$. 
The weight vector, $w$, that is learned from a training set, $\{(x_1,y_1),\dots,(x_n,y_n)\}$,
by solving the following convex optimization problem:
\begin{align}
		\min_w&\quad \frac{\lambda}{2} \| w \|^2 + \sum_{i=1}^n H_i(w),	\label{eq:main}
\end{align}
where $\lambda\geq 0$ is a regularization parameter. 
$H_i(w)$ is the (scaled) \emph{structured hinge loss} that is defined as 
\begin{align}
	H_i(w) &\!=\! \frac{1}{n}\!\max_{y \in \mathcal{Y}}\!\big\{
	\Delta(y_i, y) \!-\! \langle w, \phi(x_i, y_i) \!-\! \phi (x_i, y) \rangle\big\},  \!\!\label{eq:maxoracle}
\end{align}
where $\Delta:\mathcal{Y}\times\mathcal{Y}\to\mathbb{R}$ is a task-specific loss 
function, for example the Hamming loss for image segmentation tasks. 

Computing the value of $H_i(w)$, or the label that realizes this value, 
requires solving an optimization problem over the label set. 
We refer to the procedure to do so as the \emph{max-oracle}, or just \emph{oracle}. 
Other names for this in the literature are \emph{loss-augmented inference}, or just \emph{the $\argmax$ step}.
It depends on the problem at hand how the max-oracle is implemented. 
We give details of three choices and their properties in Appendix~\ref{sec:OracleDetails}.

Structural SVMs have proved useful for numerous complex computer vision tasks, including 
\emph{human pose estimation}~\cite{ionescu2009structural,yang2010recognizing},
\emph{semantic image segmentation}~\cite{bertelli2011kernelized,nowozin2010parameter,szummer2008learning}, 
\emph{scene reconstruction}~\cite{gupta2010estimating,schwing2012efficient} and 
\emph{tracking}~\cite{hare2011struck,lou2011structured}. 
In this work, we concentrate not on the question if SSVMs learn better predictors 
than other methods, but we study Equation~\eqref{eq:main} from the point of a 
challenging optimization problem. 
We are interested in the question how fast for given data and parameters we 
can find the optimal, or a close-to-optimal, solution vector $w$. 
This is a question of high practical relevance, since training structured SVMs 
is known to be computationally costly, especially in a computer vision context 
where the output set, $\mathcal{Y}$, is large and the max-oracle requires solving
a combinatoric optimization problem~\cite{nowozin2010parameter}.
%
%

\subsection{Related Work}
Many algorithms have been proposed to solve the optimization problem~\eqref{eq:main}
or equivalent formulations. 
In~\cite{tsochan} and~\cite{taskar}, where the problem was 
originally introduced, the authors derive a quadratic program (QP) 
that is equivalent to~\eqref{eq:main} but resembles the SVM optimization 
problem with slack variables and a large number of linear constraints.

The QP can be solved by a cutting-plane algorithm that alternates 
between calling the max-oracle once for each training example and 
solving a QP with a subset of constraints (cutting planes) obtained 
from the oracle. 
The algorithm was proved to reach a solution $\epsilon$-close to 
the optimal one within $O(\frac{1}{\epsilon^2})$ step, \ie 
$O(\frac{n}{\epsilon^2})$ calls to the max-oracle. 
Joachims \etal improved this bound in~\cite{Joachims2009} 
by introducing the \emph{one-slack} formulation.
It is also based on finding cutting planes, but keeps their 
number much smaller, achieving an improved convergence rate 
of $O(\frac{n}{\epsilon})$.
The same convergence rate can also be achieved using 
\emph{bundle methods}~\cite{teo2010bundle}.

Ratliff \etal observed in ~\cite{ratliff2007online} that one
can also apply the \emph{subgradient method} 
directly to the objective~\eqref{eq:main}, which also allows 
for stochastic and online training.
A drawback of this is that the speed of convergence depends crucially 
on the choice of a learning rate, which makes subgradient-based 
SSVM training often less appealing for practical tasks. 

The Frank-Wolfe algorithm (FW)~\cite{FWolfe56} is an elegant alternative: 
it resembles subgradient methods in the simplicity of its updates, 
but does not require a manual selection of the step size.
Recently, Lacoste-Julien \etal introduced a block-coordinate 
variant of the Frank-Wolfe algorithm (BCFW)~\cite{BCFW}.
It achieves higher efficiency than the original FW algorithm by 
exploiting the fact that the SSVM objective can be decomposed 
additively into $n$ terms, each of which is structured. 
In their experiments, the authors showed a significant speedup 
of the BCFW algorithm compared to the original FW algorithm as 
well as the previously proposed techniques.

BCFW can be considered the current state-of-the-art for SSVM training. 
However, in the next section we show that it can be significantly 
improved upon for computer vision tasks, in which the training 
time is dominated by calls to the max-oracle.


\section{Efficient SSVM Training}\label{sec:method}
In this section we introduce our main contribution, the \emph{multi-plane block-coordinate Frank-Wolfe} 
(MP-BCFW) algorithm for SSVM training.
Because it builds on top of the FW and BCFW methods, we start by giving a more detailed 
explanation of the working mechanisms of these two algorithms. 
Afterwards, we highlight the improvements we make to tackle the situation when the 
max-oracle is computationally very costly.

First, we rewrite the structured Hinge loss term~\eqref{eq:maxoracle} 
 more compactly as
\begin{equation}
	H_i(w)=\max_{y\in \calY} \; \langle \varphi^{iy},[w \; 1]\rangle, \label{eq:oracle}
\end{equation}
where $\langle\,\cdot,\cdot\,\rangle$ denotes the inner product in $\mathbb{R}^{d+1}$,
and $[w\; 1]$ is the concatenation of $w$ with a single $1$ entry.
For a vector  $\varphi\in\mathbb R^{d+1}$ we denote its first $d$ 
components as $\varphi_\star\in\mathbb R^d$ and its last component 
as $\varphi_\circ\in\mathbb R$.
The data vector $\varphi^{iy}$ in~\eqref{eq:oracle}
for $i=1,\dots,n$ and $y\in\mathcal{Y}$ is given by $\varphi^{iy}_\star=\frac{1}{n}(\phi(x_i,y)-\phi (x_i, y_i))$ and 
$\varphi^{iy}_\circ=\frac{1}{n}\Delta(y^i,y)$. 
Note that $\langle \varphi^{iy},[w \; 1]\rangle=\langle \varphi^{iy}_\star,w\rangle + \varphi^{iy}_\circ$. 

\subsection{Frank-Wolfe algorithm}
The Frank-Wolfe (FW) algorithm solves the SSVM training problem in its dual form.
Writing $H(w)=\sum_{i=1}^n H_i(w)$ and introducing concatenated vectors 
$\varphi^{\bar y}=\varphi^{(y_1,\ldots,y_n)}=\sum_{i=1}^n \varphi^{iy_i}$,
the primal problem becomes
\begin{equation}
	\min_w\; \frac{\lambda}2 ||w||^2\!+\!H(w),\ \ \text{for}\ 
	H(w)=\max_{\bar y\in \overline{\calY}} \; \langle \varphi^{\bar y},[w\,1]\rangle,
	\label{eq:single}
\end{equation}
for $\overline{\calY}=\calY\times\cdots\times\calY$. 
Note that evaluating $H(w)$ for a given $w$ requires $n$ calls to the 
max-oracle, one for each of the terms $H_1(w),\ldots,H_n(w)$.

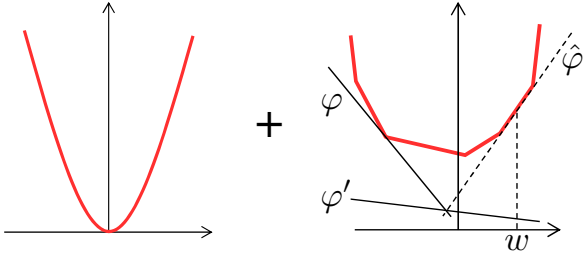
\begin{figure}[t]
\begin{center}
	\scalebox{1.3}{
		\begin{tabular}{ccc} 
			\includegraphics[height=70pt]{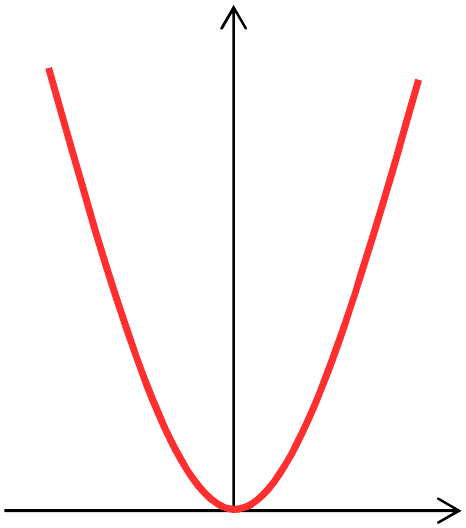} &
			\raisebox{30pt}{\Large +} &
			\includegraphics[height=70pt]{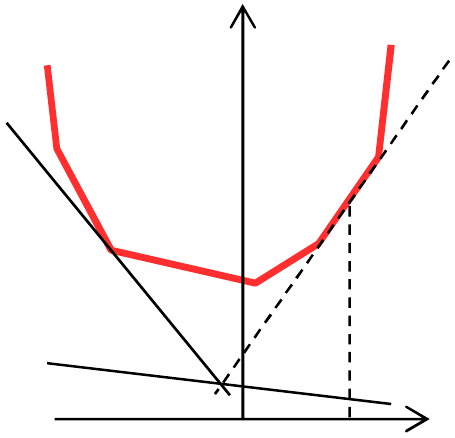}
			\begin{picture}(0, 0)
				\put(-79,38){ $\varphi$} 
				\put(-79,10){ $\varphi'$} 
				\put(-9,50){ $\hat \varphi$} 
				\put(-25.5,-3){ $w$} 
			\end{picture}
		\end{tabular}
	}
	\end{center}
	\caption{Frank-Wolfe algorithm for minimizing $\frac{\lambda}2||w||^2+H(w)$. The vector $\varphi$ 
    specifies the current linear lower bound on $H$. 
	One iteration involves three steps: (i)~Compute the vector $w$ that minimizes $\frac{\lambda}2 ||w||^2+ \langle \varphi,[w \; 1]\rangle$.
	(ii)~Obtain a new linear bound $\hat \varphi$ on $H(\cdot)$ by computing a subgradient of $H$ at $w$.
(iii)~Compute a linear interpolation $\varphi'$ between $\varphi$ and $\hat \varphi$ that maximizes $\mathcal{F}(\varphi')$, and set $\varphi\leftarrow \varphi'$.  }
\label{fig:FW}
\end{figure}

The FW algorithm maintains a (hyper)plane specified by a 
vector $\varphi\in \mathbb R^{d+1}$ that corresponds to a 
lower bound on $H(\cdot)$: $\langle \varphi,[w\; 1]\rangle\le H(w)$ 
for all $w\in\mathbb R^d$.
Such a plane exists, because $H$ is convex. In fact, any 
$\varphi^{\bar y}$ for $\bar y\in\overline{\calY}$ has this property, 
as well as any convex combination of such planes, 
$\varphi=\sum_{\bar y\in \overline\calY}\alpha_{\bar y} \varphi^{\bar y}$ 
for $\sum_{\bar y\in\overline\calY} \alpha_{\bar y}=1$ and $\alpha_{\bar y}\ge 0$.
We call vectors $\varphi$ of this form \emph{feasible}. 
%

Any feasible $\varphi=[\varphi_\star\; \varphi_\circ]$ provides a lower 
bound on~\eqref{eq:single}, which we can evaluate analytically
\begin{equation}
	\mathcal{F}(\varphi) = \min_w \Big\{\frac{\lambda}2 ||w||^2 + \langle \varphi,[w \, 1]\rangle\,\Big\}=-\frac{1}{2\lambda}||\varphi_\star||^2 + \varphi_\circ.
	\label{eq:Phi}
\end{equation}

Maximizing $\mathcal{F}(\varphi)$ over all feasible vectors $\varphi$ yields the 
tightest possible bound. This maximization problem is the dual to~\eqref{eq:single}. 
The Frank-Wolfe algorithm is iterative procedure for maximizing $\mathcal{F}(\varphi)$. 
It is stated in pseudo code in Algorithm~\ref{alg:FW}, and it is illustrated in Figure~\ref{fig:FW}.
Each iteration monotonically increases $\mathcal{F}(\varphi)$ (unless the maximum is reached),
and the algorithm converges to an optimal solution with the 
rate $O(\frac{1}{\epsilon})$ with respect to the number of iterations,
\ie $O(\frac{n}{\epsilon})$ oracle calls~\cite{BCFW}.

\begin{algorithm}[t]\small
	\caption{Frank-Wolfe algorithm for the dual of~\eqref{eq:single}
}\label{alg:FW}
\begin{algorithmic}[1]
	\STATE set $\varphi\leftarrow \varphi^{\bar y}$ for some $\bar y\in\overline\calY$
	\STATE {\bf repeat} 
	\STATE~~compute $w\leftarrow \argmin_w \frac{\lambda}{2}||w||^2 + \langle \varphi,[w\; 1]\rangle$;
	
	~~the solution is given by $w= - \frac{1}{\lambda} \varphi_\star$
	\STATE~~call oracle for vector $w$: compute $\hat \varphi\leftarrow\argmax\limits_{\varphi^{\bar y}:\bar y\in\overline\calY} \langle \varphi^{\bar y},[w\; 1]\rangle$
	\STATE~~compute $\gamma\leftarrow \argmax_{\gamma\in[0,1]} \mathcal{F}((1-\gamma)\varphi + \gamma \hat \varphi)$ as follows: \\
		~~~~~~set 
		    $\gamma\leftarrow\frac{\langle \varphi_\star-\hat \varphi_\star,\varphi_\star\rangle-\lambda (\varphi_\circ-\hat \varphi_\circ)}{||\varphi_\star-\hat \varphi_\star||^2}$
			and clip $\gamma$ to $[0,1]$ \\
	~~set $\varphi\leftarrow (1-\gamma)\varphi + \gamma \hat \varphi$
	\STATE {\bf until} some stopping criterion
\end{algorithmic}
\end{algorithm}

\subsection{Block-coordinate Frank-Wolfe algorithm}
The \emph{block-coordinate Frank-Wolfe algorithm}~\cite{BCFW} 
also solves the dual of problem~\eqref{eq:main}, but it improves 
over the FW algorithm by making use of the additive structure of the 
objective \eqref{eq:single}.
Instead of keeping a single plane, $\varphi$, it maintains $n$ planes, 
$\varphi^1,\ldots,\varphi^n$, such that the $i$-th plane is a lower 
bound on $H_i$:
$\langle \varphi^i,[w\; 1]\rangle\le H_i(w)$ for all $w\in\mathbb R^d$. 
Each such plane is obtained as a convex combination of the planes that 
define $H_i(\cdot)$, \ie $\varphi^i=\sum_{y\in \calY}\alpha_{iy} \varphi^{yi}$ 
where $\sum_{y\in\calY} \alpha_{iy}=1$ and $\alpha_{iy}\ge 0$.

We now call a vector $(\varphi^1,\ldots,\varphi^n)$ \emph{feasible} 
if each $\varphi^i$ is as above.
%
The sum $\varphi=\sum_{i=1}^n \varphi^i$ then defines a plane that 
lower bounds $H(w)=\sum_{i=1}^n H_i(w)$, \ie $\langle \varphi,[w\; 1]\rangle\le H(w)$ for all $w\in\mathbb R^d$.
Therefore, $\mathcal{F}(\varphi)$, as defined by~\eqref{eq:Phi}, is 
again a lower bound on problem~\eqref{eq:main}, 
and the goal is again to maximize this bound over all feasible vectors $(\varphi^1,\ldots,\varphi^n)$.

BCFW does so by the block-coordinate strategy. It picks an index $i\in[n]=\{1,\ldots,n\}$ 
and updates the component $\varphi^i$ while keeping all other components fixed.
During this step the terms $H_j(w)$ for $j\ne i$ are approximated by 
linear functions $\langle \varphi^j,[w \; 1]\rangle$,
and the algorithm tries to find a new linear approximation for $H_i(\cdot)$ 
that gives a larger bound $\mathcal{F}(\sum_{j=1}^n \varphi^j)$.
Pseudo code for BCFW is given in Algorithm~\ref{alg:BCFW}. 

\subsection{Multi-Plane Block-Coordinate Frank-Wolfe}
It has been shown in~\cite{BCFW} that for training SSVMs, BCFW needs much 
fewer passes through the training data than the FW algorithm as well as 
earlier approaches, such as the cutting plane and stochastic subgradient 
methods. 
However, it still has one suboptimal feature that can be improved upon:  
the computation efforts in each BCFW step are very unbalanced. 
For each oracle call (line 6 in Alg.~\ref{alg:BCFW}) there is only $\Theta(d)$ 
amount of additional work (lines 5,7,8), and this is often negligible 
compared to the time taken by the oracle.
We could easily afford to do more work per oracle call without significantly 
changing the running time of one iteration.
Our goal is therefore to exploit this extra freedom to accelerate convergence, 
thereby decreasing the number of required oracle calls and the total runtime
of the algorithm.

\begin{figure*}[t]
	\begin{center}
	\scalebox{1.3}{
		\begin{tabular}{ccccccc} 
			\includegraphics[height=70pt]{w2.pdf} &
			\raisebox{30pt}{\Large +} &
			\includegraphics[height=70pt]{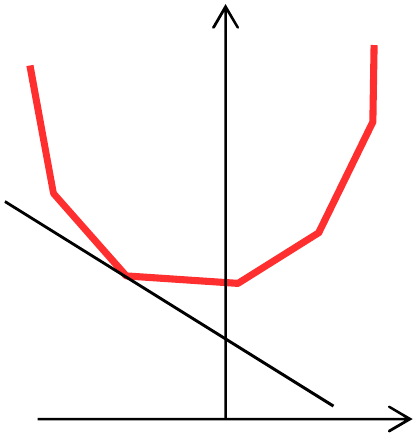} 
			\begin{picture}(0, 0)
				\put(-75,26){ $\varphi^{\mbox{\tiny $1$}}$} 
			\end{picture} &
			\raisebox{30pt}{\Large +} &
			\includegraphics[height=70pt]{H2.pdf}
			\begin{picture}(0, 0)
				\put(-85,40){ $\varphi^{\mbox{\tiny $2$}}$} 
				\put(-81,10){ $\varphi^{\mbox{\tiny $2$}'}$} 
				\put(-10,49){ $\hat \varphi^{\mbox{\tiny $2$}}$} 
				\put(-25.5,-3){ $w$} 
			\end{picture} &
			\raisebox{30pt}{\Large +} &
			\includegraphics[height=70pt]{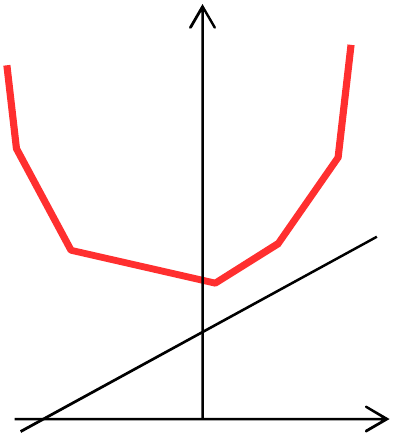} 
			\begin{picture}(0, 0)
				\put(-19,20){ $\varphi^{\mbox{\tiny $3$}}$} 
			\end{picture}
		\end{tabular}
	}
	\end{center}
	\caption{Block-coordinate Frank-Wolfe algorithm for minimizing $\frac{\lambda}2||w||^2+\sum_{i=1}^n H_i(w)$ with $n=3$.
	Each vector $\varphi^i$ specifies the current linear lower bound on $H_i$. 
    One iteration for term $i=2$ is similar to the classical Frank-Wolfe algorithm, except that 
    the terms $H_1$ and $H_3$ are approximated by linear lower bounds $\varphi^1$ and $\varphi^3$,
    respectively, which are kept fixed during this step.}
\label{fig:BCFW} 
\end{figure*}

\begin{algorithm}[t]\small
	\caption{Block-coordinate Frank-Wolfe (BCFW) \mbox{algorithm} for the dual of~\eqref{eq:main}}\label{alg:BCFW}
\begin{algorithmic}[1]
	\STATE for each $i\in[n]$ set $\varphi^i\leftarrow \varphi^{iy}$ for some $y\in\calY$

	\STATE set $\varphi=\sum_{i=1}^n \varphi^i$
	\STATE {\bf repeat}
	\STATE~~pick $i\in[n]$, \eg uniformly at random
	\STATE~~compute $w\leftarrow \argmin_w \frac{\lambda}{2}||w||^2 + \langle \varphi,[w\; 1]\rangle$
	
	~~the solution is given by $w= - \frac{1}{\lambda } \varphi_\star$
	\STATE~~call $i$-th oracle for vector $w$: \ \ $\hat \varphi^i\leftarrow\argmax\limits_{\varphi^{iy}:y\in\calY} \langle \varphi^{iy},[w\; 1]\rangle$
	\STATE~~compute $\gamma\leftarrow \argmax_{\gamma\in[0,1]} \mathcal{F}(\,\varphi-\varphi^i + (1-\gamma)\varphi^i + \gamma \hat \varphi^{i}\,)$: \\
		~~~~~~set $\gamma\leftarrow\frac{\langle \varphi^i_\star-\hat \varphi^i_\star,\varphi_\star\rangle-\lambda (\varphi^i_\circ-\hat \varphi^i_\circ)}{||\varphi^i_\star-\hat \varphi^i_\star||^2}$
			and clip $\gamma$ to $[0,1]$ 
	\STATE~~set $\varphi^{i}_{\tt old}\leftarrow \varphi^i$, ~~$\varphi^i\leftarrow (1-\gamma)\varphi^i + \gamma \hat \varphi^{i}$, ~~$\varphi\leftarrow \varphi + \varphi^i - \varphi^{i}_{\tt old}$\!\!\!\!
	\STATE {\bf until} some stopping criterion
\end{algorithmic}
\end{algorithm}

Our main insight is that BCFW acts wastefully by discarding the plane 
$\hat \varphi^i$ after completing the iteration for the term $H_i$, even 
though it required an expensive call to the max-oracle to obtain $\hat \varphi^i$. 
We propose to retain some of these planes, maintaining a \emph{working set} 
$\calW_i$  for each $i=1,\dots,n$.
We proceed similarly to~\cite{Joachims2009}, where a working set was used in
a cutting-plane framework.
Whenever the oracle for $H_i$ is called, the obtained plane $\hat \varphi^i$ 
is added to $\calW_i$.
Planes are removed again from $\calW_i$ after a certain time, unless in the 
mean time they have become \emph{active} (see below). 
Consequently, at any iteration the algorithm has access to multiple planes 
instead of just one, which is why we name the proposed algorithm 
\emph{multi-plane block-coordinate Frank-Wolfe (MP-BCFW)}.

The working set, $\calW_i$, allows us to define an alternative mechanism for 
increasing the objective value that does not require a call to the costly max-oracle. 
We define an approximation $\tilde H_i(w)$ of $H_i(w)$ by
\begin{equation}
	\tilde H_i(w)=\max\limits_{\tilde \varphi^i\in\calW_i}\langle \tilde \varphi^i,[w\; 1]\rangle.
	\label{eq:tildeH}
\end{equation}
Note that $\tilde H_i(w)\le H_i(w)$ for all $w\in \mathbb R^d$, since the maximization is performed
over a smaller set.
For any $i\in[n]$, we can perform block updates like in BCFW, but with the term $\tilde H_i$
instead of $H_i$, \ie in step 5 of Algorithm~\ref{alg:BCFW} we 
set $\hat \varphi^i\leftarrow \argmax\limits_{\tilde \varphi^i\in\calW_i}\langle\tilde \varphi^i,[w\; 1]\rangle$.
Such approximate oracle steps will increase $\mathcal{F}(\varphi)$, 
but potentially less so than a BCFW update using the exact 
expression.\footnote{Note that the function $\langle \varphi^i,[w\; 1]\rangle$ 
may not be a lower bound on $\tilde H(w)$: the plane $\varphi^i$ is a convex 
combination of planes $\{\varphi^{iy}\::\:y\in \calY\}$, but some of 
these planes may have been removed from the working set $\calW_i$. 
However, this property is not required for the correctness of the method.
Indeed, it follows from the construction that in Algorithm~\ref{alg:BCFW-WS} 
the vector $\varphi^i$ is a always convex combination of planes $\{\varphi^{iy}\::\:y\in \calY\}$ for each $i$,
and each step is guaranteed not to decrease the bound $\mathcal{F}(\sum_{i=1}^n \varphi^i)$.
%
As a consequence, the convergence properties proven in~\cite{BCFW} still hold.}

We propose to interleave the approximate updates steps with 
exact updates. 
The order of operations in our current implementation is shown in Algorithm~\ref{alg:BCFW-WS}. 
We refer to one pass through the data in lines 5 and 6 as an {\em exact} and an {\em approximate pass}
respectively, and to steps 5-6 as an {\em (outer) iteration}. Thus, each iteration
contains 1 exact pass and up to $M$ approximate passes.
The parameter $N$ bounds the number of stored planes per term: $|\calW_i|\le N$ for each $i$.
In this algorithm a plane is considered {\em active} at a given moment
if is it returned as optimal by either an exact or an approximate oracle call. 

The complexity of one approximate update in step 4 is $O(Nd)$,
therefore the algorithm performs $O(MNd)$ additional work per each 
oracle call. 
For $M=N=0$, MP-BCFW reduces to the classical BCFW algorithm.
Since MP-BCFW in particular performs all steps that BCFW does, 
it inherits all convergence guarantees from the earlier algorithm, 
such as a convergence rate of $O(\frac{1}{\epsilon})$ towards the 
optimum, as well as the guarantee of convergence even when the 
max-oracle can solve the problem only approximately (see~\cite{BCFW} for details). 
However, as we will show in Section~\ref{sec:experiments}, MP-BCFW 
gets more ``work done'' per iteration and therefore converges faster 
with respect to the number of max-oracle calls. 

\begin{algorithm}[t]\small
	\caption{Multi-Plane Block-Coordinate Frank-Wolfe (MP-BCFW) algorithm. Parameters: $N,M,T$.
	}\label{alg:BCFW-WS}
	\begin{algorithmic}[1]
		\STATE for each $i\in[n]$ set $\varphi^i\leftarrow \varphi^{iy}$ for some $y\in\calY$, 
		
		\STATE set $\varphi=\sum_{i=1}^n \varphi^i$ \\ 
		\STATE {\bf if } $N>0$~ {\bf then } $\calW_i=\{\varphi^i\}$ {\bf else } $\calW_i=\varnothing$ for each $i\in[n]$ 
		\STATE {\bf repeat} until some stopping criterion
		\STATE~~~do one pass through $i\in[n]$ in random order, \\
		~~~for each $i$ do the following: \\
			~~~~~~~run BCFW update using original term $H_i(w)$ \\
			~~~~~~~add obtained vector $\hat \varphi^i$ to $\calW_i$ \\
			~~~~~~~if $|\calW_i|\!>\!N$ then remove longest inactive plane from $\calW_i$ 
		\STATE~~do up to $M$ passes (see text) through $i\!\in\![n]$ in random order, \\
		~~for each $i$ do the following: \\
			~~~~~~~run BCFW update with term $\tilde H_i(w)\!=\!\max\limits_{\tilde \varphi^i\in\calW_i}\langle \tilde \varphi^i,[w\; 1]\rangle$\\
            ~~~~~~~remove planes from $\calW_i$ that have not been active during \\
            ~~~~~~~the last $T$ outer iterations
		\STATE {\bf end repeat}
	\end{algorithmic}
\end{algorithm}

\subsection{Automatic Parameter Selection}\label{subsec:parameter}
The optimal number of planes to keep per term as well as the 
optimal number of efficient approximate passes to run depends 
on several factors, such as the number of support vectors and how 
far the current solution still is from the optimal one. 
Therefore, we propose not to set these parameters to fixed values 
but to adapt them dynamically over time in a data-dependent way.
The first criterion described below is fairly standard~\cite{Joachims2009},
but the second criterion is specific to MP-BCFW and forms a 
second contribution of this work.

\medskip\noindent\textbf{Working set size.} \quad 
We observe that the working set is bounded not only by its upper bound 
parameter, $N$, but also by the mechanism that automatically removes 
inactive planes. 
Since the second effect is more interpretable, we suggest to 
set $N$ to a large value, and rely on the parameter $T$ to 
control the working set size.
In effect, the actual number of planes is adjusted in a 
data-dependent way for each training instance.
In particular, for terms with few relevant planes 
(support vectors) the working set will be small. 
A side effect of this is an acceleration of the algorithm, 
since the runtime of the approximate oracle is proportional 
to the actual number of planes in their working set.

\medskip\noindent\textbf{Number of approximate passes.} \quad 
We set the maximal number of approximate passes per iteration, $M$, 
to a large value and rely on the following geometrically 
motivated criterion instead.
After each approximate pass we compute two quantities: 1) the 
increase in $\mathcal{F}(\varphi)$ per time unit 
(\ie the difference of function values divided by the runtime) 
of the most recent approximate pass, and 2) the increase in 
$\mathcal{F}(\varphi)$ per unit time of the complete sequence 
of steps since starting the current iteration.
If the former value is smaller than the latter, we stop making 
approximate passes and start a new iteration with an exact pass.

\begin{figure}\centering
\includegraphics[height=.33\columnwidth]{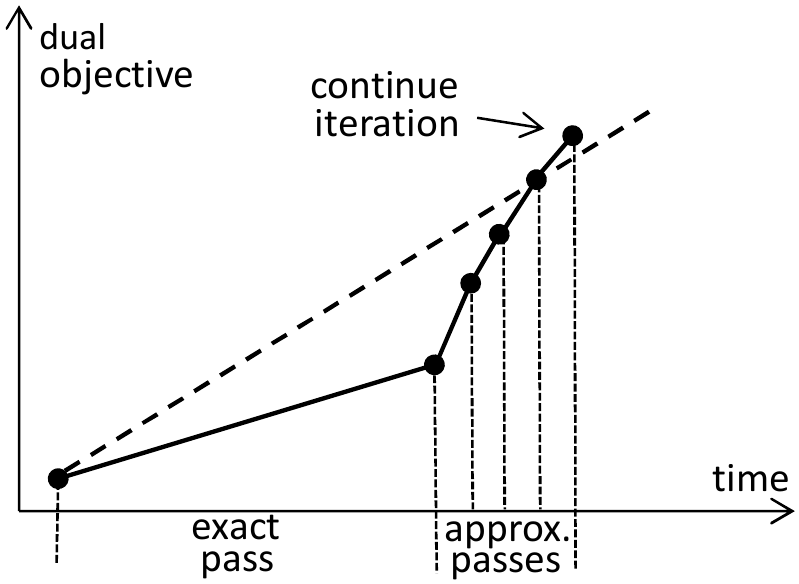}\quad
\includegraphics[height=.33\columnwidth]{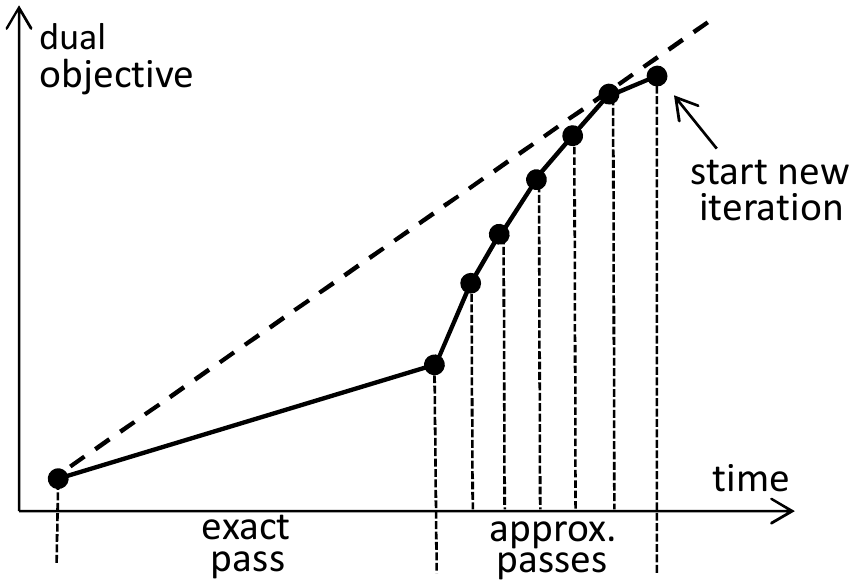}
\caption{Geometric criterion for number of approximate passes. 
After each approximate pass, we compare the relative progress (increase in objective 
per time $=$ slope of the last black line segment) to the relative progress of the complete 
iteration so far (slope of the dashed line). If the former is higher (left), we make another approximate pass. 
Otherwise, we start a new iteration (right).}\label{fig:criterion}
\end{figure}

The above criterion can be understood as an extrapolation of 
the recent behavior of the runtime-vs-function value graph 
into the future, see Figure~\ref{fig:criterion}.
If the slope of the last segment is higher than the slope of 
the current iteration so far, then the  expected increase from 
another approximate pass is high enough to justify its cost. 
Otherwise, it is more promising to start a new iteration. 

\subsection{Weighted averaging of iterates}
It has been observed that the convergence speed of stochastic 
optimization methods can often be improved further by taking 
weighted averages of iterates~\cite{bottou2012stochastic}. 
Writing $\varphi^{(k)}$ for the vector produced after the $k$-th  
oracle call in Algorithm~\ref{alg:BCFW} (BCFW), the $k$-th averaged 
vector is $\bar \varphi^{(k)}=\frac{2}{k(k+1)}\sum_{t=1}^k t \varphi^{(t)}$.
It can also be computed incrementally via  $\bar \varphi^{(k+1)}=\frac{k}{k+2}\bar \varphi^{(k)}+\frac{2}{k+2}\varphi^{(k+1)}$ for $k\ge 1$.
Several studies, including~\cite{BCFW}, has shown that the primal objective 
of the iterates $\bar w^{(k)}=-\frac{1}{\lambda}\bar\varphi^{(k)}_\star$ 
often converges to the optimum significantly faster than that of the iterates 
$w^{(k)}=-\frac{1}{\lambda}\varphi^{(k)}_\star$.

For inclusion into MP-BCFW we extended the above scheme by maintaining two vectors, $\bar \varphi^{(k)}$ 
and $\bar{\bar \varphi}^{(k')}$. They are updated after every exact and after every approximate oracle 
call, respectively, using the above formula. 
When we need to extract a solution, we compute the interpolation between $\bar \varphi^{(k)}$ and $\bar{\bar \varphi}^{(k')}$
that gives the best dual objective score.
By this construction we overcome the problem that the two types of oracle calls 
have quite different characteristics, and thus may require different weights.

We refer to the averaged variants of BCFW and MP-BCFW as BCFW-avg and MP-BCFW-avg, respectively.


\section{Experiments}\label{sec:experiments}

We analyze the effectiveness of the MP-BCFW algorithm by 
performing experiments for four different setting: multi-class
classification, sequence labeling, figure-ground segmentation
and semantic image segmentation.
The first two rely on generic datasets that were used previously 
to benchmark SSVM training (multi-class classification on the USPS 
dataset\footnote{\url{http://www-i6.informatik.rwth-aachen.de/~keysers/usps.html}} 
and sequence labeling on the OCR dataset\footnote{\url{http://www.seas.upenn.edu/~taskar/ocr/}}). 
The third task (figure-ground segmentation on part of the HorseSeg 
dataset\footnote{\url{http://www.ist.ac.at/~akolesnikov/HDSeg/}}) 
and the fourth (multiclass semantic image segmentation on the Stanford background
dataset\footnote{\url{http://dags.stanford.edu/projects/scenedataset.html}}) 
show a typical feature of computer vision tasks: the max-oracle 
is computationally costly, much more so than in the previous two 
cases, which results in a strong computational bottleneck for 
training.
In the case of multiclass segmentation the max-oracle is even NP-hard 
and can be solved only approximately, another common property for 
computer vision problems. 
Exact details of dataset characteristics, feature representations 
and implementation of the oracles for the four datasets are provided in Appendix~\ref{sec:OracleDetails}.

\begin{figure*}[!t]\centering
\subfigure[USPS]{\includegraphics[width=\textwidth]{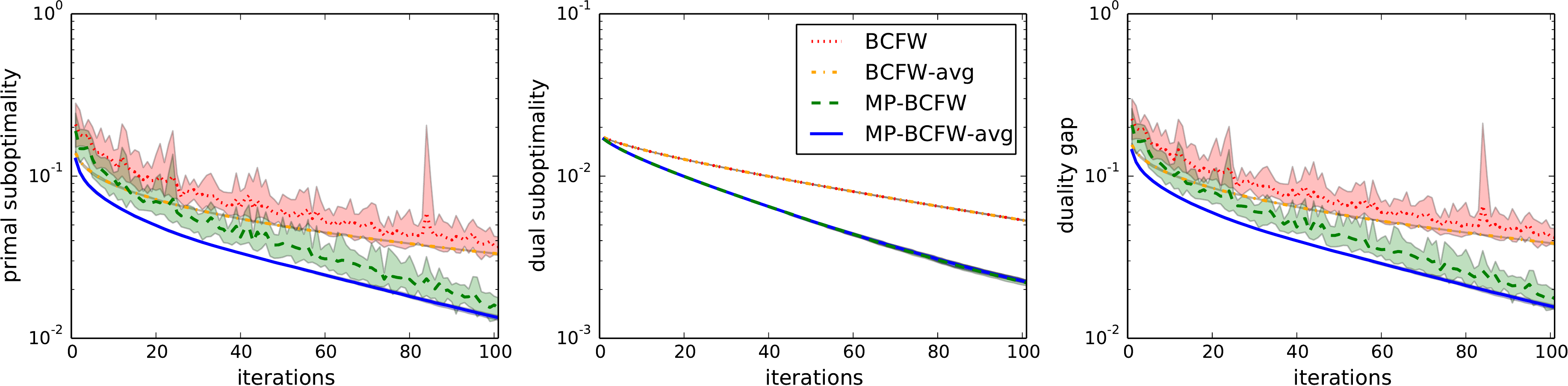}}

\subfigure[OCR]{\includegraphics[width=\textwidth]{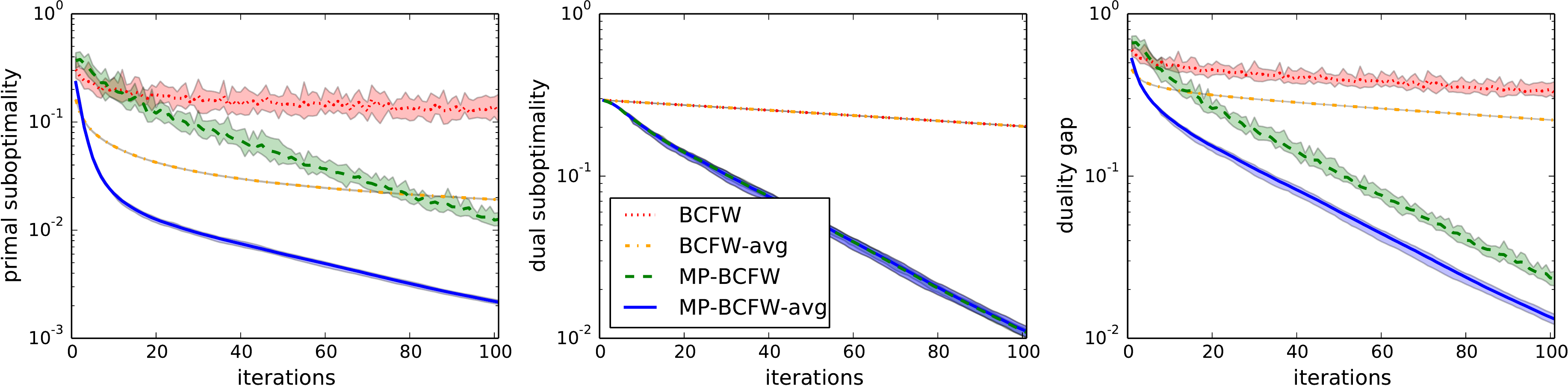}}

\subfigure[HorseSeg]{\includegraphics[width=\textwidth]{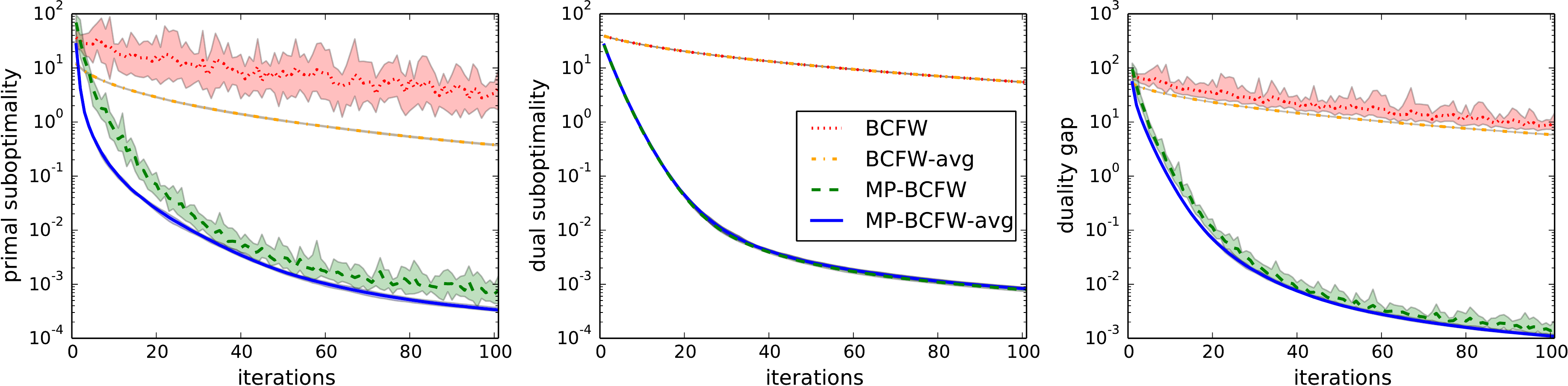}}

\subfigure[Stanford]{\includegraphics[width=\textwidth]{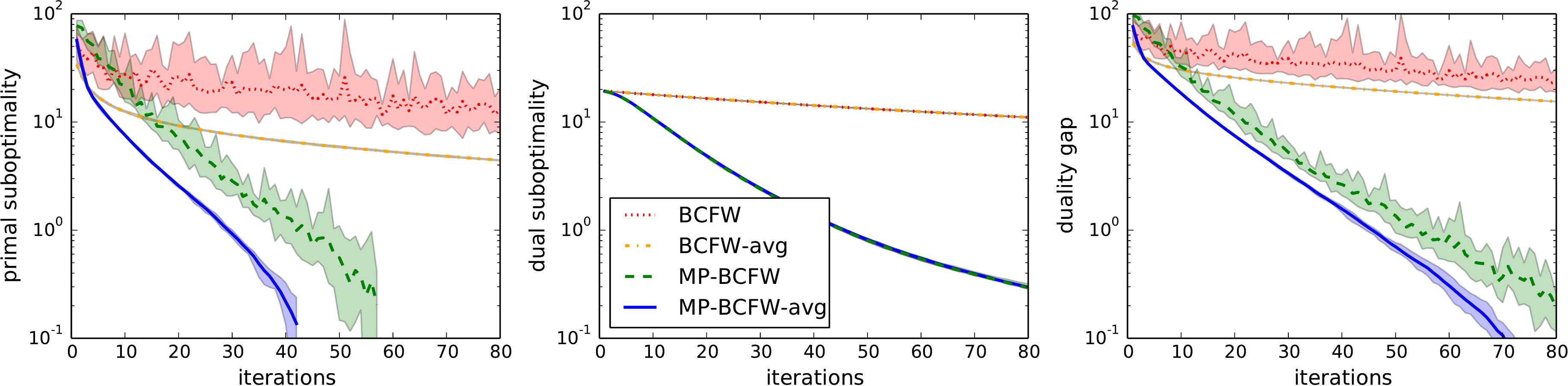}}

	\caption{\textit{Oracle convergence} on the four benchmark datasets: 
	primal suboptimality (left), dual suboptimality (middle), duality gap (right). 
	Shaded areas indicate minimum and maximum values over 10 repeats. 
    In all cases, the multi-plane algorithms, MP-BCFW and MP-BCFW-avg, 
    require fewer iterations to reach high quality solutions than the 
    single plane variants, BCFW and BCFW-avg.}
    \label{fig:oracleconv}
\end{figure*}


\begin{figure*}[!t]\centering
\subfigure[USPS]{\includegraphics[width=\textwidth]{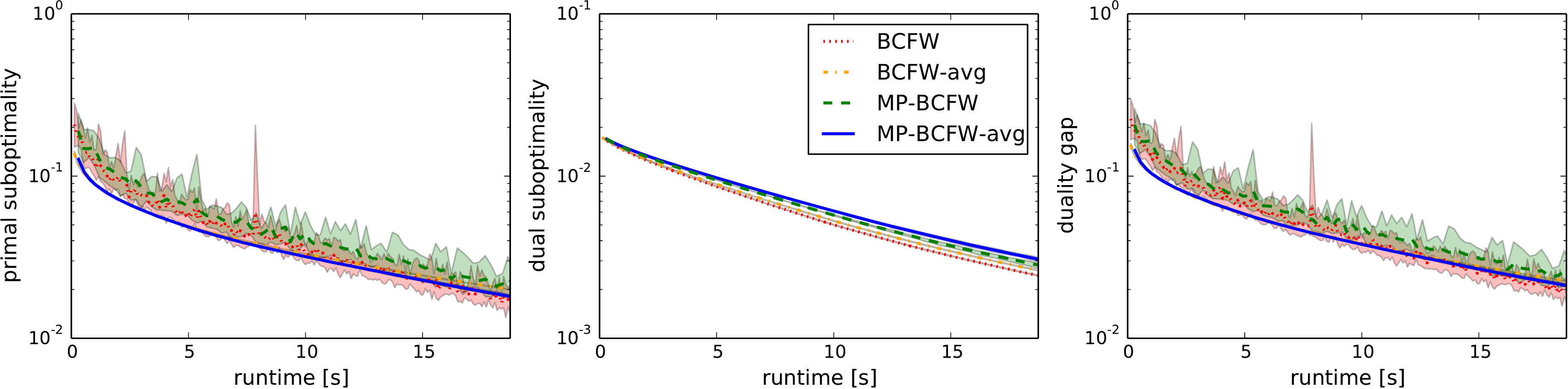}}

\subfigure[OCR]{\includegraphics[width=\textwidth]{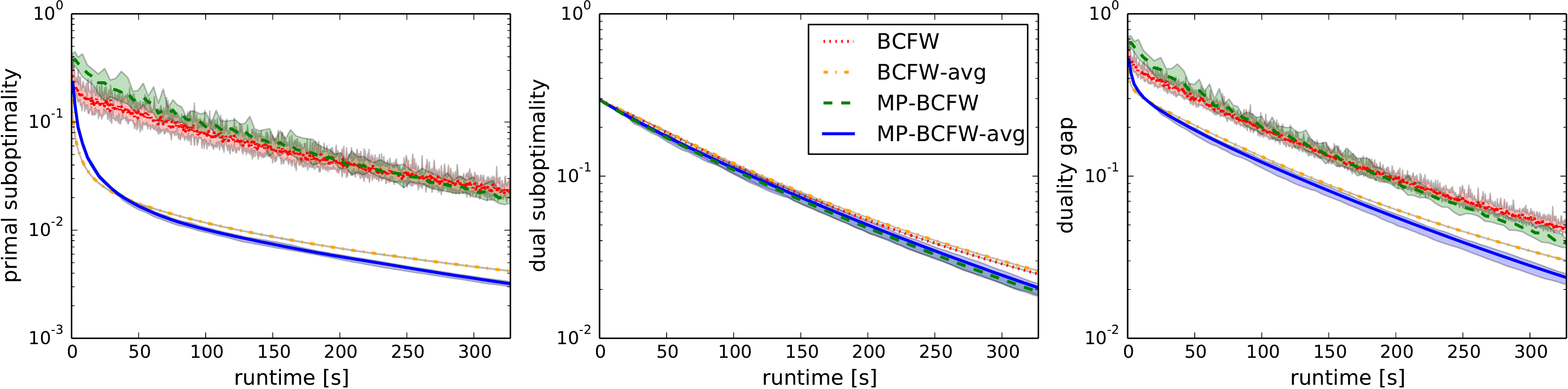}}

\subfigure[HorseSeg]{\includegraphics[width=\textwidth]{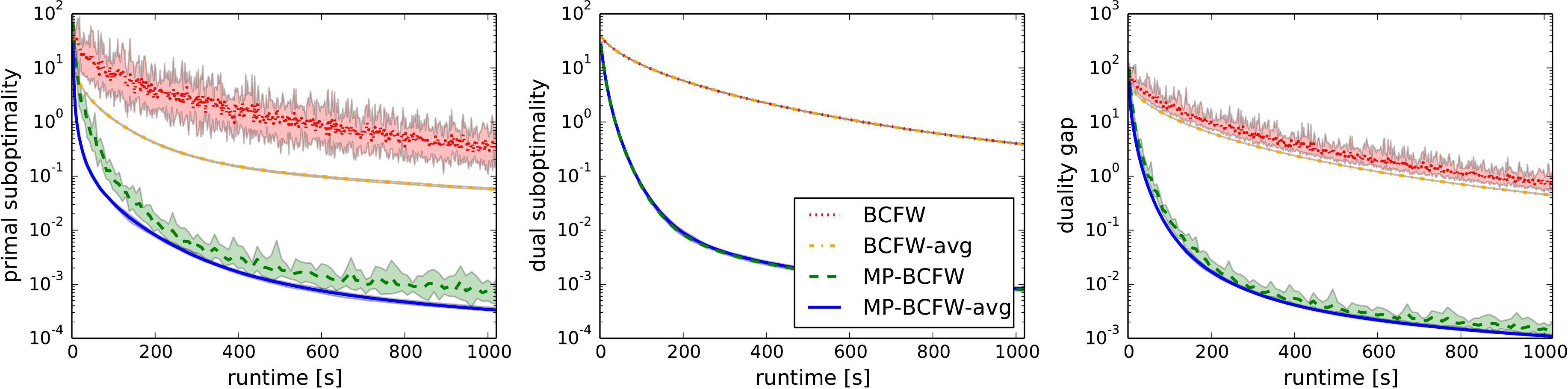}}

\subfigure[Stanford]{\includegraphics[width=\textwidth]{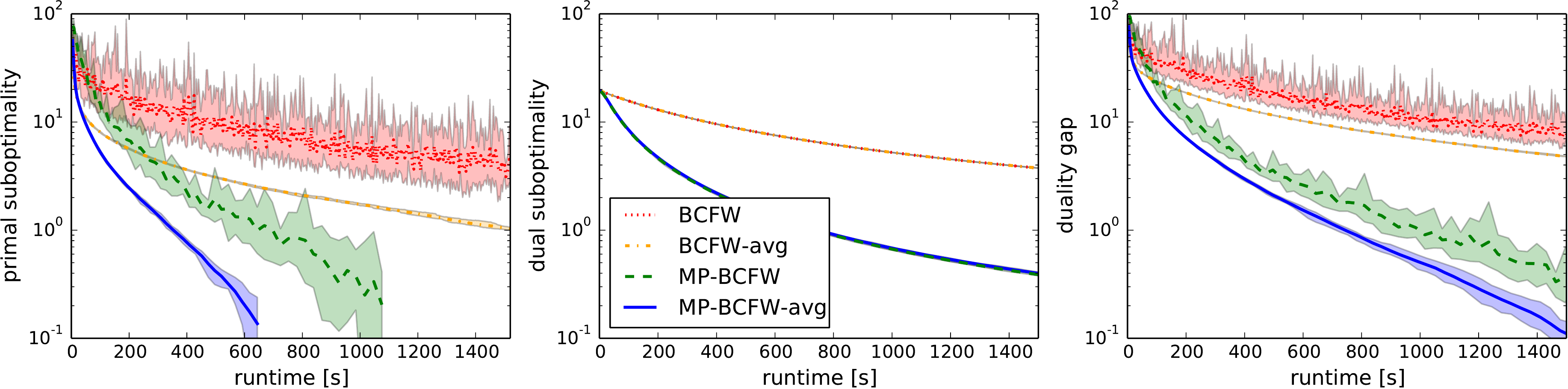}}

	\caption{\textit{Runtime convergence} on the four benchmark datasets: 
	primal suboptimality (left), dual suboptimality (middle), duality gap (right). 
	Shaded areas indicate minimum and maximum values over 10 repeats. 
    When the $\max$-oracle is fast (USPS and OCR), the multi-plane algorithms
    (MP-BCFW, MP-BCFW-avg) behave similarly to their single-plane counterparts
    (BCFW, BCFW-avg) due to the automatic parameter adjustment. 
    When the $\max$-oracle is computationally costly (HorseSeg, Stanford) 
    the multi-plane variants converge substantially faster.}
    \label{fig:runtimeconv}
\end{figure*}


We focus on comparing MP-BCFW with BCFW (with and without averaging), 
since in~\cite{BCFW} it has already be shown that  BCFW offers a 
substantial improvements over earlier algorithms, in particular 
classical FW~\cite{FWolfe56}, cutting plane  
training~\cite{Joachims2009}, exponentiated gradient~\cite{collins2008exponentiated} 
and stochastic subgradient training~\cite{ratliff2007online}. 
Since all algorithms solve the same convex optimization problem 
and will ultimately arrive at the same solution, we are only 
interested in the convergence speed, not in the error rates of 
the resulting predictors. 
This allows us to adopt an easy experimental setup in which 
we use all available training data for learning and do not 
have to set aside data for performing model selection. 
In line with earlier work, we regularize using $\lambda=1/n$. 
%
%

For all algorithms we measure three quantities: i) the 
\emph{primal suboptimality}, \ie the difference between 
the primal objective and the highest lower bound we observe 
during any of our experiments, ii) the \emph{dual suboptimality}, 
\ie the difference between the dual objective and the lowest 
observed upper bound, and iii) the \emph{duality gap}, \ie the 
difference between the primal and the dual objective.
%
%
Note that for the Stanford dataset, the reported values are 
only approximate, since evaluating the objective~\eqref{eq:main} 
exactly is intractable. In particular, the primal value is not 
guaranteed to be an upper bound on the optimal value, so the 
approximate suboptimalities and duality gap can become negative 
(at which point we interrupt the logarithmic plots). 

We visualize the results in two coordinate frames: i) with 
respect to the number of iterations, and ii) with respect to 
the actual runtime. 
The first quantity, which we refer to as \emph{oracle convergence}, 
measures how efficiently the algorithm uses the statistical information 
that is present in the training examples. 
It is independent of the implementation and therefore comparable 
between publications.
The second value, called \emph{runtime convergence}, is of practical 
interest, because it reflects the computational resources required 
to achieve a certain solution quality. 
However, it depends on the concrete implementation and computing 
hardware.\footnote{Our C++ implementation is available at \cite{codeURL}.
All experiments were performed on a desktop PC with 3.6GHz Intel Core i7 CPU. 
} 
To nevertheless obtain fair runtime comparisons, we use the same code base 
for all methods, making use of the fact that BCFW can be recovered from 
MP-BCFW with minimal overhead by deactivating the working sets and approximate 
passes ($N=0$, $M=0$). 
For MP-BCFW we rely on the automatic parameter selection mechanism
and set $T=10$, $N=1000$, $M=1000$, where the latter two just act as
high upper bounds that do not influence the system's behaviour.

\subsection{Results}
Figure~\ref{fig:oracleconv} shows the \emph{oracle convergence} 
results for all datasets. One can see that within the 
same number of iterations (and thereby calls to the max-oracle), 
MP-BCFW always achieves lower primal suboptimality than BCFW. 
Similarly, MP-BCFW-avg improves over BCFW-avg.

This effect is stronger for OCR, HorseSeg and Stanford than for 
USPS, which makes sense, since the latter dataset has a very 
small label space ($|\mathcal{Y}|=10$), so the number of 
support vectors per example is small, which limits the benefits 
of having access to more than one plane.
The graph labeling tasks OCR, HorseSeg and Stanford have a larger 
label spaces, so one can expect more support vectors to contribute 
to the score. 
Reusing planes from previous iterations can be expected to have 
a beneficial effect. 

Figure~\ref{fig:runtimeconv} illustrates the \emph{runtime convergence},
\ie the values on the vertical axis are identical to Figure~\ref{fig:oracleconv}, 
but the horizontal axis shows the actual runtime instead of the number of 
iterations.
One can see that for the USPS dataset, the better oracle convergence 
did not translate to actually faster convergence, and for OCR, the 
difference between single-plane and multi-plane methods is small. 

The situation is different for the HorseSeg and Stanford datasets,
which are more typical computer vision tasks.
For them, MP-BCFW and MP-BCFW-avg converge substantially faster 
than BCFW and BCFW-avg.
The differences can be explained by the characteristics of the 
max-oracle in the different optimization problems: for USPS and 
OCR these are efficient and not do not form major computational 
bottlenecks. 
For USPS, the max-oracle requires only computing ten inner products 
and identifying their maximum. This takes only a few microseconds on 
modern hardware. 
Overall, the BCFW algorithm spends approximately 15\% of its 
total runtime on oracle calls. 
For OCR, the max-oracle is implemented efficiently via Dynamic Programming (Viterbi algorithm),
which takes approximate 50{\textmu}s on our hardware. 
Overall, oracle calls make up for approximately 70\% of BCFW's runtime. 
In both settings, the ratio of time spent for the oracle calls 
and the time spent elsewhere is not high enough to justify 
frequent use of the approximate oracle.

\begin{figure*}[t]
\centering
		\includegraphics[width=.24\textwidth]{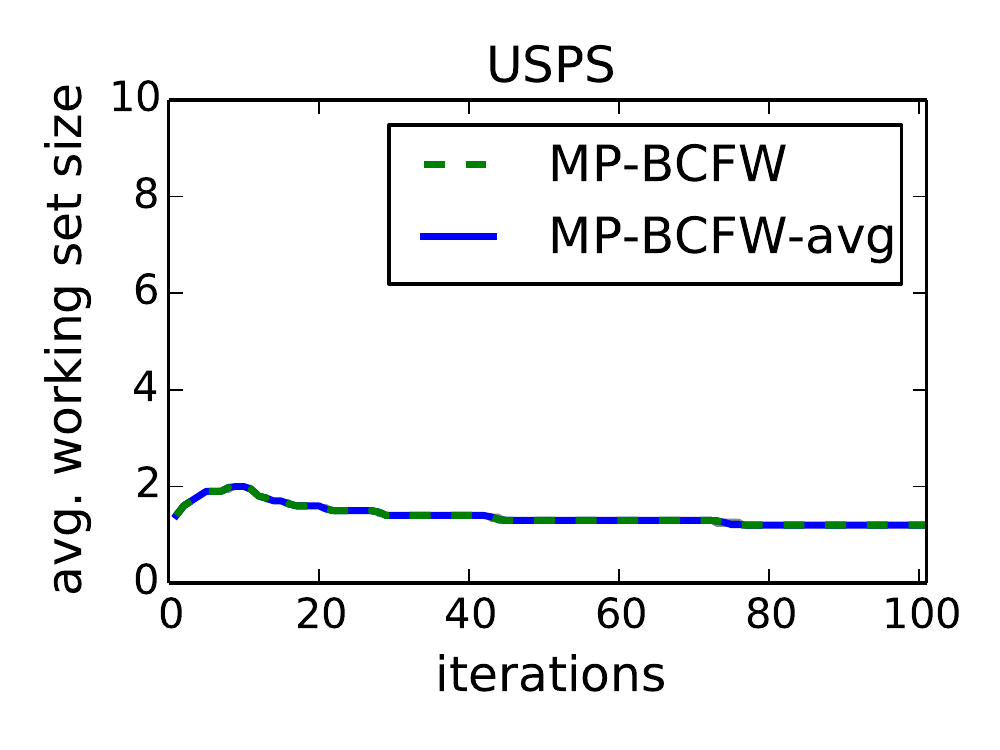}	
		\includegraphics[width=.24\textwidth]{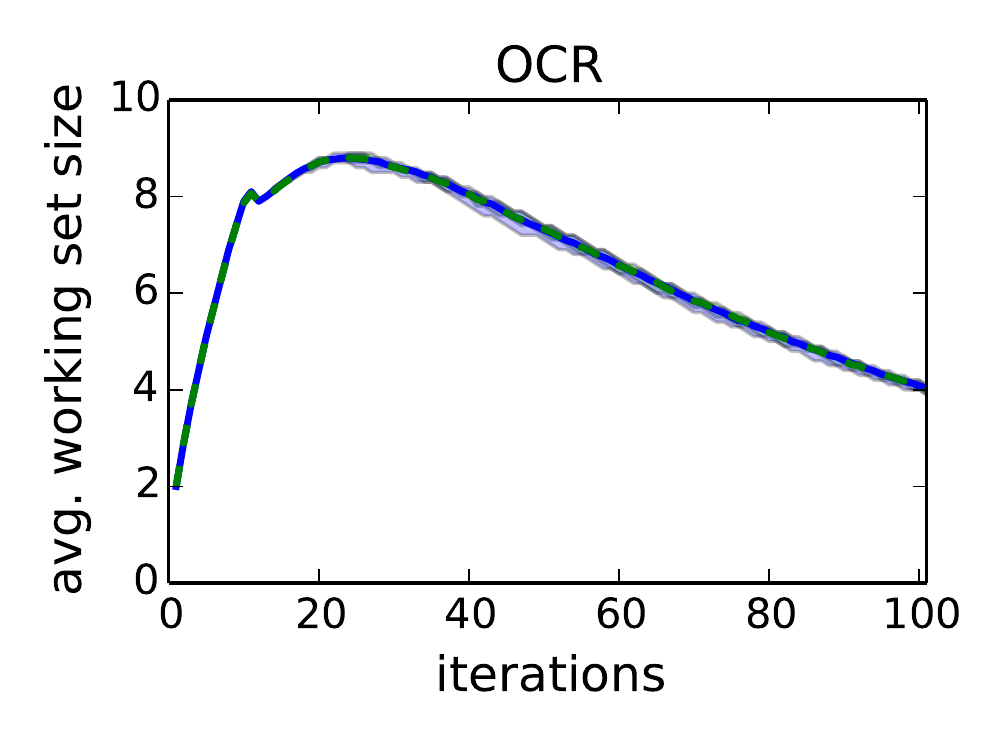}
		\includegraphics[width=.24\textwidth]{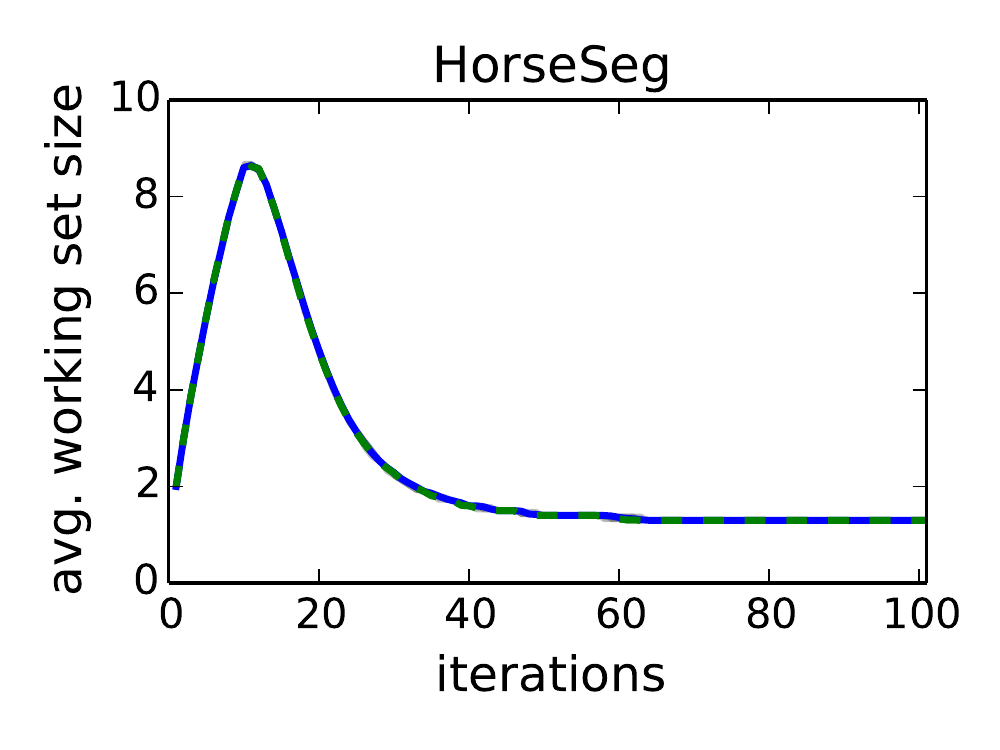}	
		\includegraphics[width=.24\textwidth]{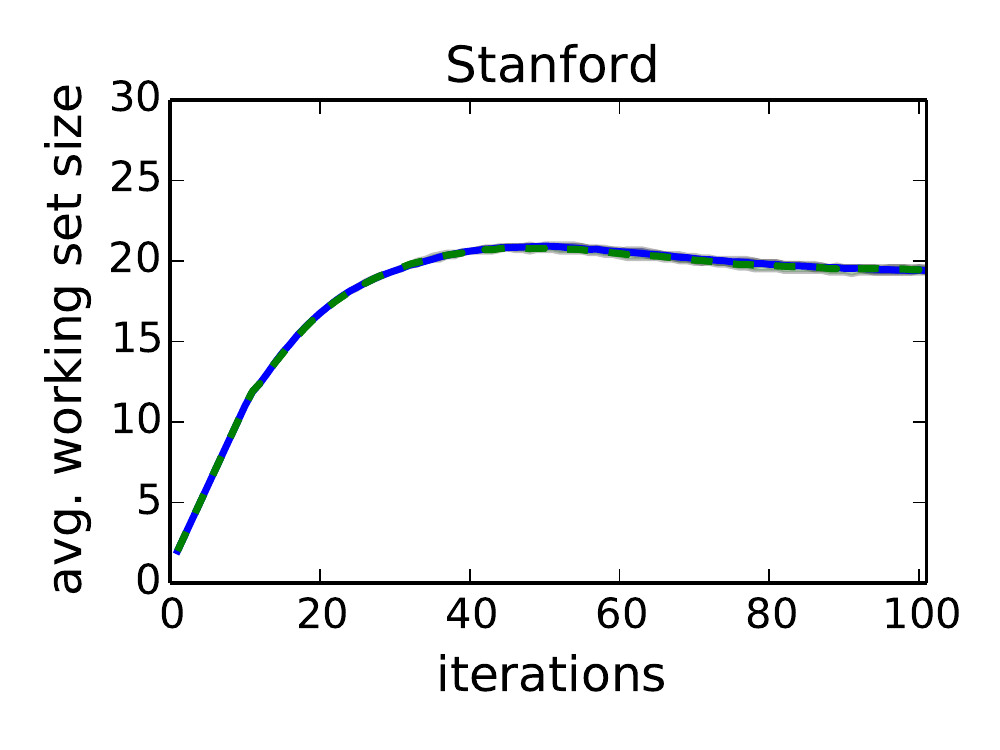}
		\\
		\includegraphics[width=.24\textwidth]{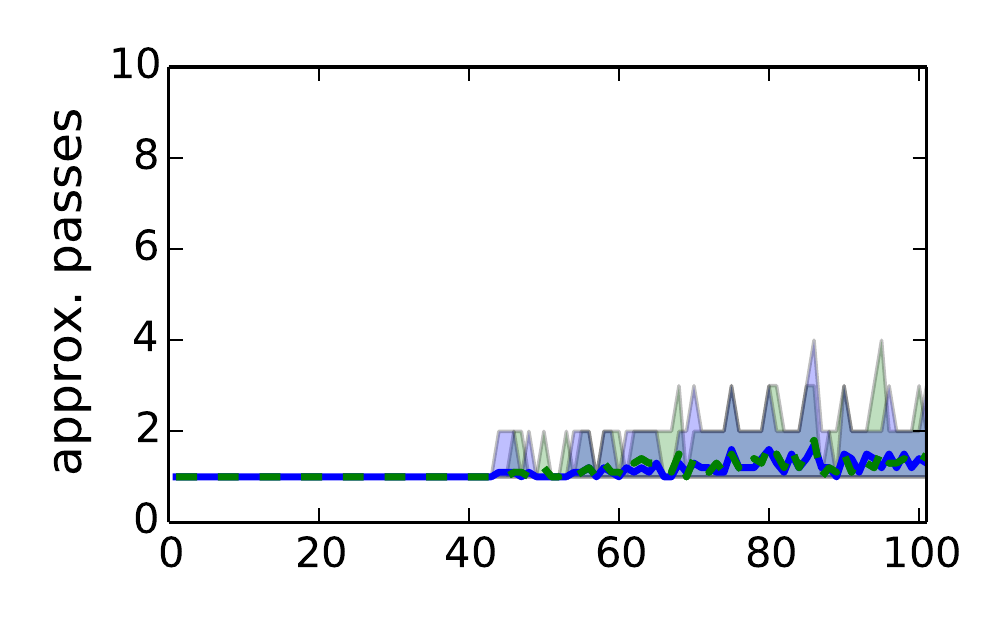}	
		\includegraphics[width=.24\textwidth]{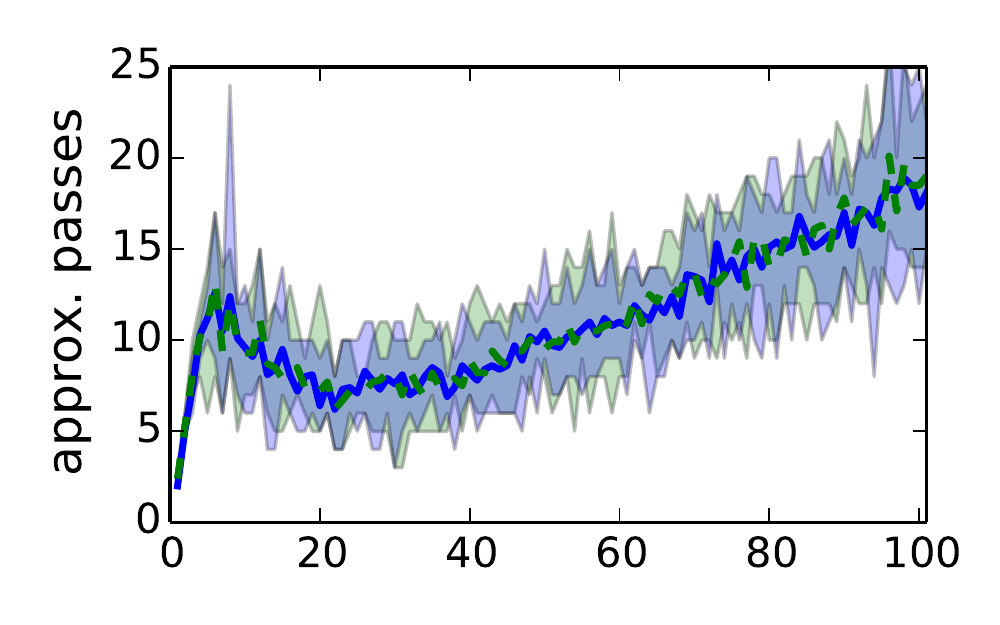}
		\includegraphics[width=.24\textwidth]{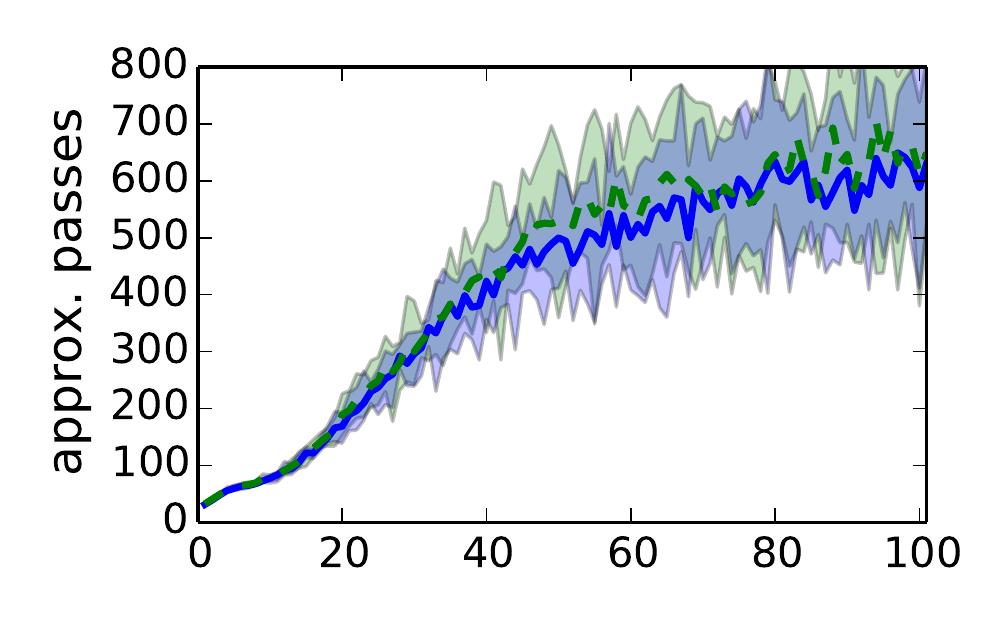}	
		\includegraphics[width=.24\textwidth]{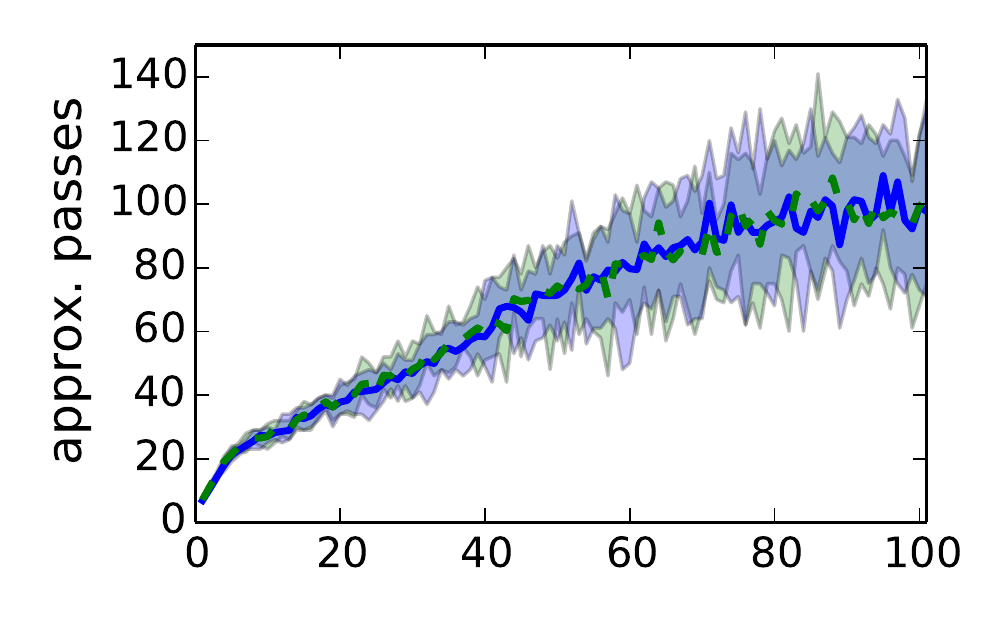}
	\caption{\textit{Automatic parameter selection.} The size of the working set (top) and the number of approximate passes per 
	iteration (bottom) are adjusted in a data- and runtime-dependent way. See Section~\ref{subsec:parameters} for details.}
	\label{fig:parameters}
\end{figure*}

Note, however, that even for USPS, MP-BCFW is not slower than BCFW, 
either. 
This indicates that the automatic parameter selection does its job 
as intended: 
if the overhead of a large working set and many approximate passes 
would be larger than the benefit they offer, the selection rule 
falls back to the established (and efficient) baseline behavior. 

For the other two datasets, the max-oracle calls are strong 
computational bottlenecks.
For HorseSeg, each oracle call consists of minimizing a 
submodular energy function by means of solving a maximum 
flow problem~\cite{boykov2004experimental}. 
For the Stanford dataset, the oracle consists of minimizing 
a \emph{Potts}-type energy function 
for which we rely on an efficient preprocessing in~\cite{gridchyn2013potts} followed by the approximation algorithm in~\cite{BVZ:PAMI01}.
Even with optimized implementations, 
the oracle calls take 1 to 5 milliseconds, and as a consequence, 
BCFW spends almost 99\% of the total training time on oracle calls.
Using MP-BCFW, the fraction drops to less than 25\%, because the parameter 
selection mechanism decides that the time it takes to make one exact oracle 
call is often spent better for making several approximate oracle calls.

\subsection{Automatic parameter selection}\label{subsec:parameters} 

Figure~\ref{fig:parameters} illustrates the dynamic behavior 
of the parameter selection for MP-BCFW in more detail. 
The first row shows the average size of the working set per 
term over the course of the optimization, the second row shows 
the number of approximate passes per iteration.

One can see that for USPS the number of planes and iterations 
remain small through the runtime of the algorithm, making MP-BCFW 
behave almost as BCFW. 
For OCR, the algorithm initially identifies several relevant 
planes, but this number is reduced for later iterations. 
The number of approximate passes per iteration ranges 
between 10 to 15. 

For HorseSeg shows a similar pattern but in more extreme form. 
After an initial exploration phase the number of relevant planes
stabilizes at a low value, while the number of approximate passes 
grows quickly to several hundred. 
For Stanford, the working set grows steadily to reach a stable state,
but the number of approximate passes grows slowly during the course
of the optimization.

In summary, one can see that the automatic parameter selection shows a 
highly dynamic behaviour. It allows MP-BCFW to adapt to the complexity of 
the objective function as well as to dynamic properties of the optimization, 
such as how far from the objective the current solution is.


\section{Summary and Conclusion}\label{sec:summary}
We have presented MP-BCFW, a new algorithm for training structured SVMs 
that is specifically designed for the situation of computationally costly 
max-oracles, as they are common in computer vision tasks.
The main improvement is the option to re-use previously computed oracle 
results by means of a per-example working set, and to alternate between 
calls to the exact but slow oracle and calls to an approximate but fast oracle. 
We also introduce rules for dynamically choosing the working set size and
number of approximate passes depending on the algorithm's runtime behaviour 
and its progress towards the optimum. 
Overall, the result is a easy-to-use method with default settings that 
work well for a range of different scenarios.
Our C++ implementation is publicly available at~\cite{codeURL}.

Our experiments showed that MP-BCFW always requires fewer iterations 
to reach a certain solution quality than the BCFW algorithm, which 
had previously been shown to be superior to earlier methods. 
This leads to faster convergence towards the optimum for situations 
in which the max oracle is the computational bottleneck, as it 
is typical for structured learning tasks in computer vision. 

In future work we plan to explore the situation of very high-dimensional 
or kernelized SSVMs, where a further acceleration can be expected by 
precomputing and caching inner product evaluations.


\section*{Acknowledgements}
We thank Alexander Kolesnikov for his help with the HorseSeg dataset.
This work was funded by the European Research
Council under the European Unions Seventh Framework Programme (FP7/2007-2013)/ERC grant agreement no 308036
and 616160.

\appendix

\section{Details of Datasets and max-Oracles}\label{sec:OracleDetails}

\noindent We give details of the four different applications scenarios with max-oracles of increasing complexity.


\subsection*{Multiclass Classification -- USPS Dataset}
The USPS\footnote{\url{http://www-i6.informatik.rwth-aachen.de/~keysers/usps.html}}  
dataset is a standard multiclass dataset with 7291 samples of 10 classes, $\mathcal{Y}=\{\texttt{0},\texttt{1},\dots,\texttt{9}\}$. 
Each sample, $x$, comes with a 256-dimensional feature vector, 
$\psi(x)$, from which we build a 2560-dimensional joint feature 
map 
\begin{equation}
	\phi(x,y) = \Big(\psi(x)\llbracket y=\texttt{0}\rrbracket,
	\dots, \psi(x)\llbracket y=\texttt{9}\rrbracket\Big).
\label{eq:phi}
\end{equation}
As loss function we use the ordinary multiclass loss, 
$\delta(y,\bar y)=\llbracket y\neq \bar y\rrbracket$. 
Overall, the structured Hinge loss is 
\begin{equation}
H_i(w) = \frac{1}{n} \max_{y\in\mathcal{Y}} \Big\{ \ \llbracket y_i\neq y\rrbracket + \big\langle\,w, \phi(x_i,y)-\phi(x_i,y_i)\,\big\rangle\ \Big\}
\end{equation}
Because of the small label set the max oracle can be performed efficiently by an explicit search over all labels.

\subsection*{Sequence Labeling - OCR dataset}
The OCR dataset consists of 6877 data samples that are 
sequences of hand-written letters, $x=(x^1, x^2, \dots, x^{L})$, where for 
each part, $x^l$, a feature vector $\psi(x^l)\in\mathbb{R}^{128}$ 
is available.
The outputs are sequences of labels, $y=(y^1, y^2, \dots, y^{L})$
of the same length, where $y^l\in\{\texttt{a},\dots,\texttt{z}\}$ for each
$l=1,\dots,L$. 
The length of the sequences in fact differs for different examples with an average of 7.6. 
We do not indicate this explicitly in order to keep the notation 
simple. 

We define a joint feature map, $\phi(x,y)=\big(\phi_u(x,y),\phi_p(x,y)\big)$, 
consisting of a unary and a pairwise component:
\begin{equation*}
	\phi_u(x,y) = \sum_{l=1}^L \phi(x^l,y^l),
\qquad
	\phi_p(x,y) = \sum_{l=1}^{L-1} e_{y^l,y^{l+1}},
\end{equation*}
where $\phi(x_l,y_l)$ is a multiclass encoding of the feature $\psi(x_l)$ 
as defined above for the USPS dataset, and $e_{a,b}$ is the $26^2$-dimensional 
binary vector indicating the label pair $(a,b)$ out of all possible 
label transitions. 
As loss function we use the normalized Hamming loss,
$\Delta(y,\bar y)= \frac{1}{L}\sum_{l=1}^L \llbracket y^l \neq \bar y^l \rrbracket$.

The structured Hinge loss can be written using a summation over all
positions and transitions:
\begin{eqnarray}
H_i(w) &=& \frac{1}{n} \max_{y\in\mathcal{Y}} \Big\{ 
\sum_{l=1}^L 
\frac{1}{L}\llbracket y^l_i\neq y^l\rrbracket 
\nonumber \\&&
+ \langle\,w_u, \phi(x^l_i,y^l)-\phi(x^l_i,y^l_i)\,\rangle
\nonumber \\&&
+
\sum_{l=1}^{L-1} \langle\,w_p, e_{y^l,y^{l+1}}-e_{y^l_i,y^{l+1}_i}\,\rangle\ \Big\},
\end{eqnarray}
where $w=(w_u,w_p)$ is a decomposition of the weight vector 
into parts acting on the unary and pairwise part of the joint 
feature map, accordingly.
Even though the size of the label space grows exponentially in the 
length of the sequence, $|\mathcal{Y}|=26^L$, the additive structure 
of the problem makes it possible to solve the max-oracle efficiently 
using dynamic programming, namely by the Viterbi algorithm~\cite{viterbi1967error}.

\subsection*{Image Segmentation -- HorseSeg Dataset and Stanford Backgrounds Dataset}
We use a subset of 2376 image of the HorseSeg dataset~\cite{kolesnikoveccv2014},\footnote{\url{http://www.ist.ac.at/~akolesnikov/HDSeg/}} 
and the 715 images of the Stanford Backgrounds dataset~\cite{GouldICCV09},\footnote{\url{http://dags.stanford.edu/projects/scenedataset.html}}
respectively. 
First, each image, $x$, is decomposed into a variable number of superpixels, $x^1,\dots,x^L$, 
using the SLIC method~\cite{SLIC}. 
Then, each superpixel is represented by a 649-dimensional feature vector $\phi(x^l)$. 
The task consists of predicting a segmentation, $y=(y^1,\dots,y^L)$, for each image. 
For the \emph{HorseSeg} dataset, this is a figure-ground mask, \ie each superpixel is 
assigned binary, $y^l\in\{0,1\}$ for every $l=1,\dots,L$. 
For the \emph{Stanford} dataset, each superpixel is assigned one of nine semantic 
class labels, $y^l\in\{1,\dots,9\}$ for every $l=1,\dots,L$. 

We construct a joint feature map using the same construction as for the 
unary term in the OCR dataset, 
\begin{equation*}
	\phi(x,y) = \sum_{l=1}^L \phi(x^l,y^l),
\end{equation*}
where $\phi$ is defined by Equation~\eqref{eq:phi}. 
In addition, we add a pairwise term of Potts type to the SSVM prediction function, 
\begin{equation*}
	\Theta(x,y) = \sum_{k\sim l} \llbracket y^k \neq y^l\rrbracket
\end{equation*}
where $k\sim l$ denotes that the superpixels $k$ and $l$ are neighbors
in the image. For this term, we do not learn a weight vector, but 
assign it a constant weight of $1$. 
Formally, the term therefore is not part of the feature vector but contributes to 
the $\varphi_\circ$ component of the parameterization (see Section 3 of the main 
manuscript). 

As loss function we again use the normalized Hamming loss and we obtain 
the following structured Hinge loss 
\begin{eqnarray}
H_i(w) &=& \frac{1}{n} \max_{y\in\mathcal{Y}} \Big\{ 
\sum_{l=1}^L 
\frac{1}{L}\llbracket y^l_i\neq y^l\rrbracket 
\nonumber \\ &&
+ \langle\,w_u, \phi(x^l_i,y^l)-\phi(x^l_i,y^l_i)\,\rangle
\nonumber \\ &&
+ \sum_{k\sim l} \llbracket y^k \neq y^l\rrbracket \ \Big\}.
\end{eqnarray}

For \emph{HorseSeg}, the objective in this discrete optimization is submodular, 
and we implement the max-oracle using the mincut algorithm~\cite{boykov2004experimental}.
For \emph{Stanford}, the resulting optimization function is NP-hard in general. 
As max-oracle we use an approximation given by the alpha-expansion algorithm~\cite{BVZ:PAMI01}
combined with an efficient preprocessing from~\cite{gridchyn2013potts}.

{\small
\bibliographystyle{ieee}
\bibliography{mpbcfw}
}

\end{document}